\documentclass[10pt,twocolumn,letterpaper]{article}

\usepackage[pagenumbers]{cvpr} 

\usepackage[dvipsnames]{xcolor}

\definecolor{cvprblue}{rgb}{0.21,0.49,0.74}
\usepackage[pagebackref,breaklinks,colorlinks,citecolor=cvprblue]{hyperref}

\def\paperID{5015} 
\def\confName{CVPR}
\def\confYear{2024}

\newcommand*\fw[1]{\textcolor{black}{#1}}

\title{LLM4VG: Large Language Models Evaluation for Video Grounding}

\author{Wei Feng\\
Tsinghua University\\
\and
Xin Wang\thanks{Corresponding author}\\
Tsinghua University\\
\and
Hong Chen\\
Tsinghua University\\
\and
Zeyang Zhang\\
Tsinghua University\\
\and
Houlun Chen\\
Tsinghua University\\
\and
Zihan Song\\
Tsinghua University\\
\and
Yuwei Zhou\\
Tsinghua University\\
\and
Yuekui Yang\\
Tencent\\
Tsinghua University\\
\and
Haiyang Wu\\
Tencent\\
\and
Wenwu Zhu$^*$\\
Tsinghua University\\
}
\usepackage{makecell}
\usepackage{multirow}
\begin{document}
\maketitle
\begin{abstract}
Large language models (LLMs) have achieved great success in various tasks. 
Recently, researchers have attempted to investigate the capability of LLMs in handling videos and proposed several video LLM models. However, the ability of LLMs to handle video grounding (VG), which is an important time-related video task requiring the model to precisely locate the start and end timestamps of temporal moments in videos that match the given textual queries, still remains unclear and unexplored in literature. To fill the gap, in this paper, we propose the LLM4VG benchmark, which systematically evaluates the performance of different LLMs on video grounding tasks. Based on our proposed LLM4VG, we design extensive experiments to examine two groups of video LLM models on video grounding: (i) the video LLMs trained on the text-video pairs (denoted as VidLLM), and (ii) the LLMs combined with pretrained visual description models such as the video/image captioning model. 
\fw{We propose tailored prompt methods to integrate the instruction of VG and description from different kinds of generators, including caption-based generators for direct visual description and VQA-based generators for information enhancement.} We also provide comprehensive comparisons of various VidLLMs and explore the influence of different choices of visual models, LLMs, prompt designs, etc, as well. Our experimental evaluations lead to two conclusions: (i) the existing VidLLMs are still far away from achieving satisfactory video grounding performance, and more time-related video tasks should be included to further fine-tune these models, and (ii) the combination of LLMs and visual models shows preliminary abilities for video grounding with considerable potential for improvement by resorting to more reliable models and further guidance of prompt instructions.
\end{abstract}    
\section{Introduction}
\label{sec:intro}


With the rapid development of large language models (LLMs) in recent years, various tasks beyond natural language processing, such as dynamic graphs~\cite{zhang2023llm4dyg}, visual question answering~\cite{guo2023images}, reinforcement learning~\cite{10.5555/3618408.3618558}, have begun to combine with LLMs for performance improvement. Also, more and more tasks requiring multimodal information are dedicated to linking LLMs' text processing capabilities with video perception abilities~\cite{zhang2023multimodal,ye2023mplug}. For instance, Li et al.~\cite{2023videochat} and Zhang et al.~\cite{damonlpsg2023videollama} propose large language models such as Video-Chat and Video-LLaMA that can handle video data,  demonstrating impressive ability in receiving and understanding video. 

Video grounding (VG), as an important time-related video task aiming to identify the corresponding video segments of given textual descriptions~\cite{wang2022video}, asks to precisely understand temporal boundary information with start and end time of different segments in videos~\cite{feng2023multimedia}. 
However, despite the success of the existing LLMs, their ability to handle video grounding (VG) which requires accurate localization of time boundaries for moments, still remains unclear and unexplored in literature.

To fill the gap, we propose LLM4VG, a comprehensive benchmark which systematically evaluates the performance of VG task for LLMs. 
We adopt two methods to complete video grounding. Based on our proposed LLM4VG, we examine two groups of video LLM strategies on VG task: i) the video LLMs trained on the text-video dataset directly accept the video content and video grounding task instructions as input and then output the prediction results (denoted as VidLLM); ii) the LLMs combined with pretrained visual description model that converts video content to text descriptions via visual description generators, thus bridging the visual and textual information. 
\fw{As for the second group of strategies, we specifically design prompts that integrate the instruction of VG and the given visual description information from different kinds of generators, including caption-based generators to directly output description and VQA-based to enhance the description information, which compensates for the failure of the caption model to include keywords of grounding query, revealing LLMs' temporal understanding abilities on video grounding tasks.}

We conduct extensive evaluations to analyze the performance of employing \textbf{six} visual models, \textbf{three} LLMs, and \textbf{three} prompting methods, and compare them with \textbf{three} VidLLMs which are directly instructed to conduct VG task. Furthermore, we claim \textbf{eight} experimental observations as foundations for designing good video LLMs on VG. 
Specifically, our evaluations show that VidLLMs are still far away from achieving satisfactory VG performance, and more time-related video tasks should be included to further finetune the VidLLMs in order to reach a performance boost.
In terms of the combining visual models and LLMs, our proposed strategy which incorporates LLMs with visual models achieves better performance for temporal boundary understanding than VidLLMs, showing preliminary abilities for VG task. 
The video grounding ability of combining strategy is mainly limited by the prompt designs and visual description models. 
More fine-grained visual models should be utilized so that more visual information is introduced to empower LLMs with the capability of understanding the visual scene and therefore adequately completing the VG task. 
\fw{In addition, the prompting method with further guidance of instructions is also required to help LLMs better conduct the VG task.}

To summarize, we make the following contributions:
\begin{itemize}
    \item We propose LLM4VG, the first comprehensive benchmark for evaluating LLMs on video grounding (VG) task. 
    \item \fw{We develop an integration of task instruction of VG and visual description from different kinds of visual generators, including caption-based generators and VQA-based generators, which proves to be effective LLM prompts.}
    \item We systematically evaluate and analyze the VG performance of different groups of video LLM models through combinations with different visual description models and promoting methods. 
    \item We conclude fine-grained observations about LLMs' performance on VG, which can serve as foundations for designing good video LLMs on VG. 
    


\end{itemize}




\section{Related Work}
\subsection{LLMs for Video Understanding}

Large Language Model (LLM) is trained through massive text data~\cite{radford2018improving} and is able to perform a wide range of tasks including text summarization, translation, reasoning, emotional analysis, and more~\cite{kojima2022large,zhou2022least,min2023recent}. With the introduction of GPT-3~\cite{floridi2020gpt},  InstructGPT~\cite{ouyang2022training}, and GPT-4~\cite{gpt4}, this concept has become widely known for understanding and generating human language.


However, in the digital world today, video and audio content share the same importance with textual content as part of multimedia data~\cite{zhu2020multimedia}. This makes it hard for a simple LLM to expand into the field of audiovisual perception to meet the needs of users. To address this issue, two different approaches have been proposed in the academic community. One is to develop a large language model with multimodal information processing capabilities. Therefore, many large language models that can handle videos have emerged, such as Video-LLaMA~\cite{damonlpsg2023videollama}, Video-Chat~\cite{2023videochat} and Video-ChatGPT~\cite{Maaz2023VideoChatGPT}. We collectively refer to them as VidLLM. While retaining LLM's powerful language comprehension abilities, VidLLM has also demonstrated impressive ability in receiving and understanding visual and auditory content. The other one is to convert the visual and audio information into intermediate language descriptions instead of dense vectors~\cite{aradhye2009video2text}, using the descriptions and few-shot in-context exemplars to instruct normal LLMs to complete video-related tasks, which has been used by Guo et al. for visual question answering tasks of LLMs~\cite{guo2023images}.

 Naturally, we wonder whether the method mentioned above can complete cross-modal tasks related to video, such as video grounding. 


\subsection{Video Grounding}

Video grounding is a task that requires the model to localize the starting and ending times of the segment target from a video~\cite{chen2018temporally}, which has drawn increasing attention over the past few years~\cite{lan2023survey,nan2021interventional}, since video grounding task is closely related to quite a few computer vision and language processing methods such as video understanding, video retrieval, and human-computer interaction, etc~\cite{yuan2021closer,gabeur2020multi,huang2018makes}.

Regarding the challenges of video grounding tasks,  many approaches have been proposed~\cite{xu2019multilevel,yang2022tubedetr,soldan2021vlg}. He et al. propose a reinforcement learning method that includes an adjustable temporal window with a sliding boundary, which has the learned policy for video grounding~\cite{he2019read}.  Zeng et al. proposed a dense regression network that regresses the distances from every frame to the starting or ending frame of the video segment described by the query~\cite{zeng2020dense}. Chen et al. proposed an Adaptive Dual-branch Promoted Network (ADPN) that exploits consistency and complementarity of audio and visual information instead of focusing on visual information only~\cite{chen2023curriculum}. These methods, however, all require the use of annotated specific video grounding training datasets for pre-training, which cannot be directly applied to task scenarios.

\begin{figure}
    \centering
    \includegraphics[width=\linewidth]{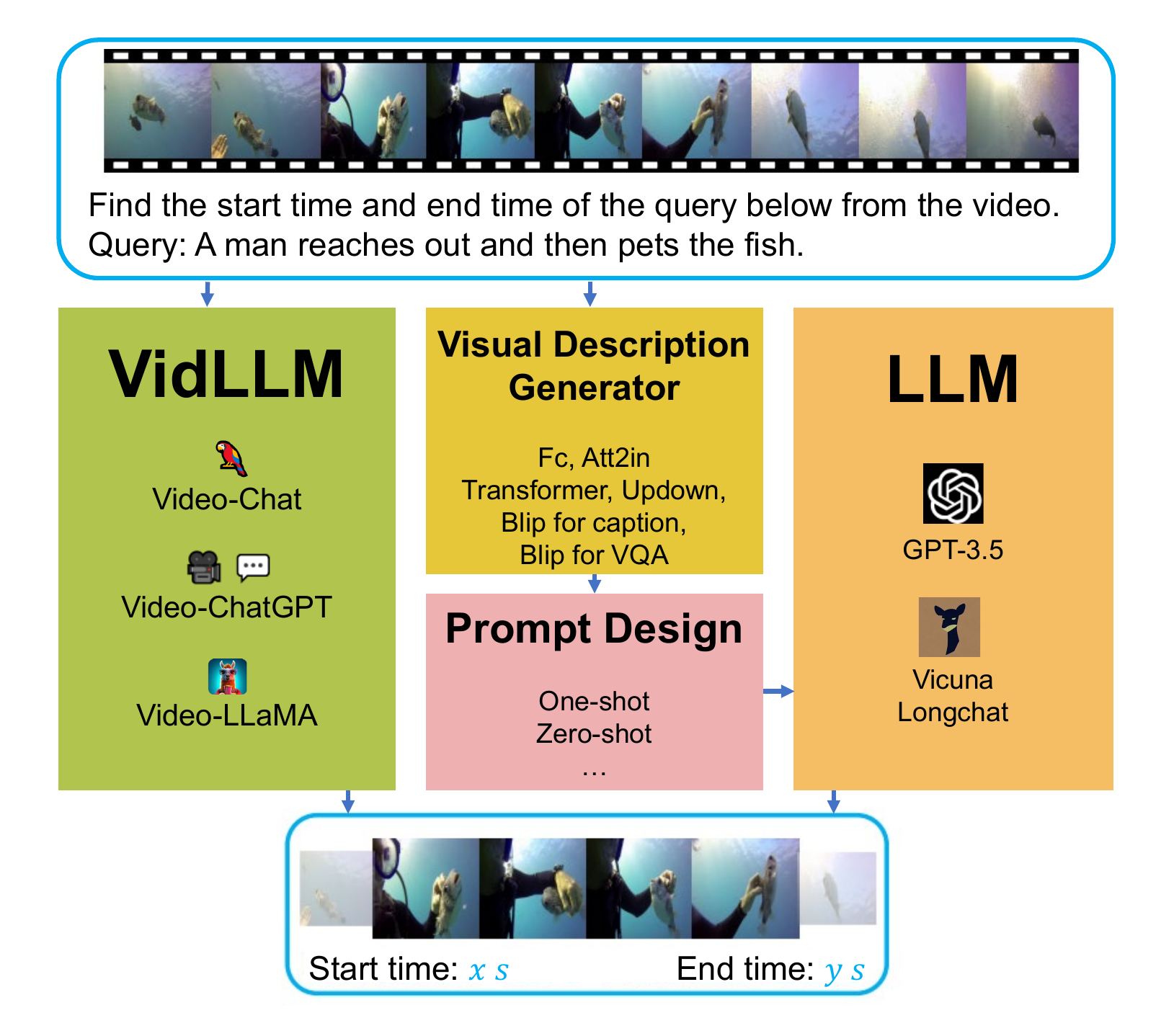}
    \caption{Benchmark of LLM4VG. We analyze the influences of applying six visual description generators, three LLMs, and three prompting methods for video grounding, comparing them with three VidLLMs which are directly instructed to conduct video grounding tasks.}
    \label{fig:benchmark}
    \vspace{-0.6cm}
\end{figure}

\begin{figure*}[h]
    \centering
    \includegraphics[width=\linewidth]{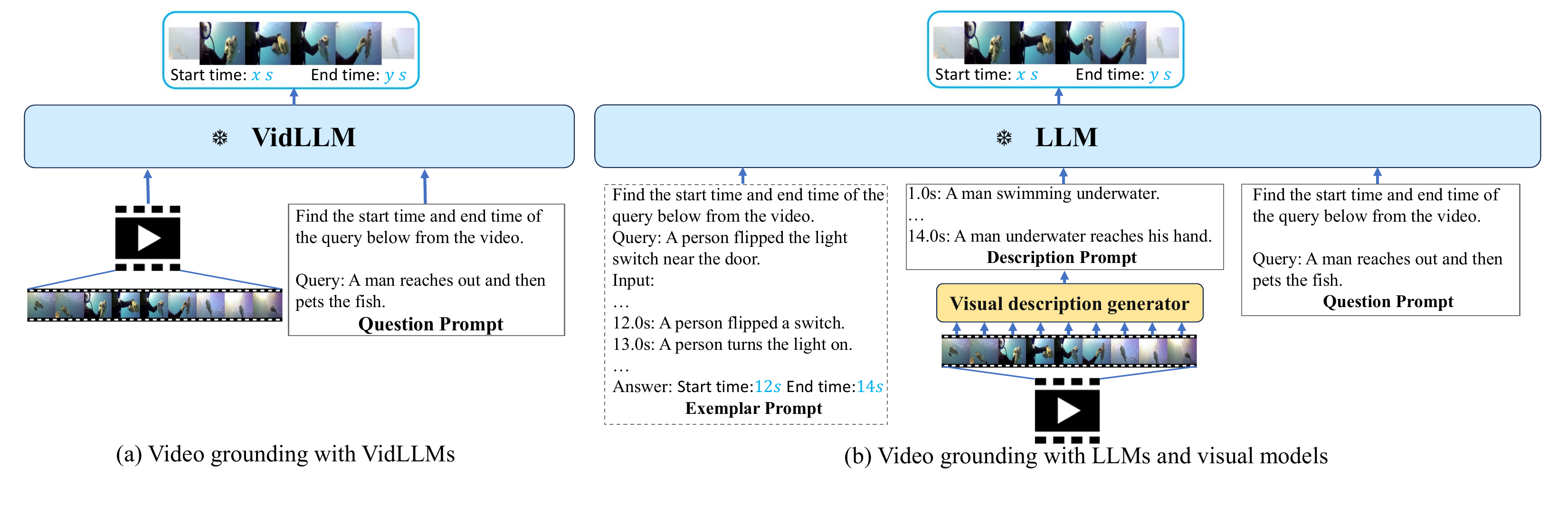}
    \vspace{-1cm}
    \caption{Framework of video grounding for LLMs. (a) stands for video grounding with VidLLMs. (b) stands for video grounding with LLMs and visual models. The dashed box represents that in the one-shot method, we will input the exemplar prompt, description prompt, and question prompt, while in the zero-shot method, we will not input the exemplar prompt.}
    \vspace{-0.6cm}
    \label{fig:framework}
\end{figure*}

\section{The LLM4VG Benchmark}

\label{sec:formatting}
In this section, we will introduce our proposed LLM4VG benchmark to evaluate whether LLMs are capable of understanding temporal information on the video grounding task. As shown in Figure~\ref{fig:benchmark}, our benchmark mainly includes four variables to be evaluated for their impact on completing video grounding tasks, including the selection of VidLLMs, normal LLMs, visual description models, and prompt designs. We will then introduce their role in completing the video grounding task in sequence.
 
\subsection{Video Grounding with VidLLMs}
As shown in Figure~\ref{fig:framework}(a), we first use VidLLMs that can access video content as the baseline of our experiment, trying to complete the video grounding task. They will directly receive video and instruct prompt to output video grounding predictions. The details of the instruct prompt are consistent with the question prompt mentioned in the following prompt design section~\ref{sec: prompt_design}.

\subsection{Video Grounding with Combination of LLMs and Visual Models}
As shown in Figure~\ref{fig:framework}(b), for those LLMs without the ability to process visual data, we first used a visual description generator to process the video, generating a basic description sequence with controllable time span parameters (such as a second-by-second caption of individual video content). Next, we adjust the description sequence to an appropriately formatted prompt as input, instructing LLM to output grounding predictions. Based on this process, we evaluate the result of video grounding from three different perspectives, which include visual description generators, prompt designs, and LLMs.


\subsubsection{Visual Description Generator}
In order to convert video data information into text content that LLMs can understand, we first extract images from the video at 1 FPS, then input images to the visual models, and then output text describing the frame at that timestamp, summarizing them to form a continuous visual description $Des = \{(t_1,c_1),(t_2,c_2)...(t_m,c_m)\}$, where $t\sim T(\{1s,2s,...\})$ is a sequential timestamp and $c_i$ is a visual description of the corresponding time. The visual models we use can be divided into caption-based and VQA-based.

\noindent\textbf{Caption-based generator.} We use a series of caption models including the simple Fc model using CNN and LSTM networks~\cite{rennie2017self,luo2021goal}, the Attention model(Att2in)~\cite{rennie2017self,luo2018discriminability} and Updown model~\cite{anderson2018bottom} introducing attention mechanisms, the transformer-based sequence modeling framework(Transformer)~\cite{li2019entangled,vaswani2017attention}, and advanced caption models such as the Blip model that effectively utilizes the noise web data by bootstrapping the captions to improve visual language task capabilities~\cite{li2022blip}. As the easiest way, these models would directly transform the image into the visual description $c_i$ per second.

\noindent\textbf{VQA-based generator.} Considering the occasional missing key information in the visual description due to the weak generalization ability of caption-based generator(for example, many visual descriptions provided by the caption model do not contain keywords in the query), we also use the Blip model with visual question answering(VQA) capability as a visual description generator~\cite{li2022blip,li-etal-2023-lavis} to enhance the description information. \fw{We first use its answer to \textit{`What is happening in the image'} as the caption description for the video at time $t_i$, and then ask it to answer \textit{`Is it currently happening $<$query event$>$ in the image'}. Finally, we will merge the two answers as the video description $c_i$ of the current time and form the description sequence $Des = \{(t_1,c_1),(t_2,c_2)...(t_m,c_m)\}$.}

\subsubsection{Prompt Design}
\label{sec: prompt_design}
To instruct the LLM for the video grounding task, we design the input text prompt for the LLM that mainly consists of three parts: question prompt, description prompt, and exemplar prompt.

The question prompt $Ques$ mainly describes the task of video grounding, which consists of task requirements and a query for video. The task requirements are \textit{`Find the start time and end time of the query below from the video'}.
The description prompt is the description sequence $Des$ received from the visual description generator, which includes the video description content of every second until the end of the video. The exemplar prompt $Exem$ is a video grounding example that we pre-generated, including the combined content of a hypothetical description prompt and a hypothetical question prompt, and an answer to it.
Finally, as shown in Figure~\ref{fig:framework}(b), we propose a zero-shot method to integrate the description prompt and question prompt as input, and we add the extra exemplar prompt to enable LLMs to better understand the task of video grounding for the one-shot method. 

Shown in Table~\ref{tab:example}, we form the prompt prepared for LLMs as input:

\begin{equation}
    Prompt=[Exem ,Des, Ques],
\end{equation}
where $Exem$ is optional depending on the chosen of one-shot or zero-shot method.
\begin{table}[!htbp]
\small
    \centering
    \begin{tabular}{p{3cm}p{4.5cm}}
    \toprule
        Prompt & Example \\
    \midrule
        Exemplar & Here is an example: Question: Given a sequence of video descriptions with the time stamps $[(t_1,c_1),(t_2,c_2)...(t_m,c_m)]$. When is the woman cooking? Answer:$[15s,21s]$ \\
    \midrule
        Description &  A sequence of video description with the time stamps $[(t_1,c_1),(t_2,c_2)...(t_m,c_m)]$.\\
    \midrule
        Question & Find the start time and end time of the query below from the video. Query: the person flipped the light switch near the door. \\
    \midrule
        Answer & $[10s,14s]$ \\
    \bottomrule
    \end{tabular}
    \caption{An example of prompt construction for the video grounding task.} 
    \vspace{-0.8cm}
    \label{tab:example}
\end{table}

\subsubsection{Large Language Model}
For LLMs that cannot directly access video content, they will be input the $Prompt$ generated in the process above and instructed to complete the video grounding task. For the result $Output = LLM(Prompt)$, we will extract the content in the answer for the prediction result of start and end time.
\subsection{Video Grounding Evaluation}
We evaluate the results of LLM in video grounding tasks. And corresponding evaluation dimension was used to measure their ability to complete video grounding. We will introduce the definition and measurement of corresponding evaluation dimensions as follows:

\textbf{Evaluation: Recall on Video Grounding}. Recall is the main outcome evaluation metric for verifying the LLMs' completion of video grounding tasks, which directly calculates the difference between the grounding time answer provided by LLMs and the actual results. In the usual process of evaluating video grounding results, we first calculate the intersection over union ratio (IoU) based on the predicted results and ground truth and then use $R@n, IoU=m$ as the evaluation metrics~\cite{gao2017tall}, which represents the percentage of testing video grounding samples that have at least one correct prediction (i.e., the IoU between the ground truth and the prediction result is larger than $m$) in the top-n results of prediction.

\begin{table*}
\small
    \centering
    \begin{tabular}{cccccccc}
    \toprule
        Valid Rate&Model&Fc~\cite{rennie2017self}&Att2in~\cite{rennie2017self}&Transformer~\cite{li2019entangled}&Updown~\cite{anderson2018bottom}&Blip~\cite{li2022blip}&Blip(VQA)~\cite{li2022blip}\\
    \midrule
    \multicolumn{8}{c}{\textit{Zero-Shot Evaluation with Large Language Model}}\\
        \multirow{3}{*}{\makecell{R@1 IoU=0.3\\Random:23.36}}
        &GPT-3.5~\cite{gpt2022}&25.83&25.99&25.46&23.74&25.81&25.97\\
        &Vicuna-7B~\cite{vicuna2023}&19.87&19.19&20.73&19.92&19.41&19.57\\
        &Longchat-7B~\cite{longchat2023}&23.47&23.17&23.01&23.90&22.55&23.95\\
        \multirow{3}{*}{\makecell{R@1 IoU=0.5\\Random:9.06}}
        &GPT-3.5&9.68&10.19&10.62&9.03&10.03&10.05\\
        &Vicuna-7B&8.20&7.72&8.04&7,72&8.06&7.90\\
        &Longchat-7B&9.38&9.60&9.84&10.91&9.60&9.95\\
        \multirow{3}{*}{\makecell{R@1 IoU=0.7\\Random:2.88}}
        &GPT-3.5&2.50&2.58&2.31&2.50&3.20&3.04\\
        &Vicuna-7B&2.53&2.55&2.42&2.12&2.47&2.23\\
        &Longchat-7B&3.09&2.82&3.25&3.47&2.82&3.33\\
    \midrule
    \multicolumn{8}{c}{\textit{One-Shot Evaluation with Large Language Model}}\\
        \multirow{3}{*}{\makecell{R@1 IoU=0.3\\Random:23.36}}
        &GPT-3.5&24.11&23.47&24.19&24.25&26.02&17.96\\
        &Vicuna-7B&16.72&16.94&17.66&17.82&15.22&16.29\\
        &Longchat-7B&19.70&19.03&18.68&18.25&19.30&19.54\\
        \multirow{3}{*}{\makecell{R@1 IoU=0.5\\Random:9.06}}
        &GPT-3.5&9.30&8.90&9.60&9.11&10.91&7.61\\
        &Vicuna-7B&7.10&7.23&7.45&7.02&5.99&7.02\\
        &Longchat-7B&8.87&8.58&8.33&7.69&8.33&8.74\\
        \multirow{3}{*}{\makecell{R@1 IoU=0.7\\Random:2.88}}
        &GPT-3.5&2.85&2.23&3.04&2.85&2.93&2.69\\
        &Vicuna-7B&2.12&2.23&1.94&1.94&1.85&2.10\\
        &Longchat-7B&2.88&2.96&2.12&2.31&2.61&2.31\\    
    \midrule
    \multicolumn{8}{c}{\textit{One-Shot+confidence judgment Evaluation with Large Language Model}}\\
        \multirow{3}{*}{\makecell{R@1 IoU=0.3\\Random:23.36}}
        &GPT-3.5&24.68&28.23&26.32&25.13&30.67&33.87\\
        &Vicuna-7B&21.67&20.70&21.67&19.30&22.37&23.63\\
        &Longchat-7B&23.20&23.71&22.82&24.41&24.57&23.63\\
        \multirow{3}{*}{\makecell{R@1 IoU=0.5\\Random:9.06}}
        &GPT-3.5&9.27&11.34&9.92&9.03&11.26&11.80\\
        &Vicuna-7B&9.76&9.09&9.38&8.74&8.20&9.14\\
        &Longchat-7B&9.60&10.22&9.01&10.22&9.97&9.76\\
        \multirow{3}{*}{\makecell{R@1 IoU=0.7\\Random:2.88}}
        &GPT-3.5&2.55&2.85&2.80&2.55&3.84&4.22\\
        &Vicuna-7B&3.15&2.77&2.69&2.80&2.45&2.63\\
        &Longchat-7B&3.17&3.17&2.45&3.92&3.31&2.96\\    

    \bottomrule
    \end{tabular}
    \caption{The overall model performance on the video grounding with different visual description generators, Large Language Models, and prompting methods. The `Blip' means we use the Blip model for captioning, while the `Blip(VQA)' means the Blip model is used for visual question answering and captioning. Considering that in some cases the visual descriptions obtained by LLMs may not be applicable to video grounding tasks, we added an extra confidence judgment prompt to check whether the description sequence is suitable for video grounding tasks. } 
    \label{tab:total}
    
    \begin{tabular}{ccccc}
        \toprule
         Valid Rate &Video-Chat~\cite{2023videochat}&Video-ChatGPT~\cite{Maaz2023VideoChatGPT}&Video-LLaMA~\cite{damonlpsg2023videollama}&Random  \\
         \midrule
         R@1 IoU=0.3&9.03&20.00&10.38&23.36\\
         R@1 IoU=0.5&3.31&7.69&3.84&9.06\\
         R@1 IoU=0.7&1.26&1.75&0.91&2.88\\
        
        \bottomrule
    \end{tabular}
    \caption{The overall model performance on the video grounding with VidLLMs} 
    \vspace{-0.4cm}
    \label{tab:VidLLM}
\end{table*}

\section{Experiments}
In this chapter, we conduct experiments to evaluate LLMs' ability to understand temporal information and language reasoning on video grounding problems.

\subsection{Setups}

\noindent\textbf{Visual description generators}.   We used caption models such as Att2in, Fc~\cite{rennie2017self,luo2018discriminability}, Transformer~\cite{li2019entangled}, Updown~\cite{anderson2018bottom}, and Blip~\cite{li-etal-2023-lavis,li2022blip} model as the main visual description generators to generate a per-second description sequence of video content. In addition, we also used the VQA model~\cite{li-etal-2023-lavis,antol2015vqa}, to generate a per-second description sequence and additional answering sequence. (i.e., answering that in every second of the video if the event mentioned in the video grounding question is happening)

\noindent\textbf{Prompts}. To investigate the impact of different prompt methods on the model's ability to complete video grounding, we compared different prompt methods, including zero-shot prompting and one-shot prompting~\cite{brown2020language,liu2023pre}. For examples shown in Table~\ref{tab:example}, the prompt is composed of exemplar prompts, description prompts, and question prompts.

\noindent\textbf{Models}.    Considering that the prompt obtained through the visual description generator generally has a large number of tokens, we used GPT-3.5-turbo-16k~\cite{gpt2022}, Vicuna-7B~\cite{vicuna2023}, Longchat-7B~\cite{longchat2023}, and VidLLM such as Video-Chat, Video-ChatGPT and Video-LLaMA~\cite{2023videochat,chen2023videollm,Maaz2023VideoChatGPT,damonlpsg2023videollama}. As a comparison, we apply a random method that randomly generates answers within the video duration for grounding. For a few prompts that LLMs refuse to provide answers due to the poor quality of visual description, the answer will be randomly generated using the rando baseline method. For VidLLM, it does not require the use of a visual description generator, as it already has the ability to receive and process video data.

\noindent\textbf{Data}.    We use the Charades-STA dataset~\cite{gao2017tall} for video grounding tasks, which is a benchmark dataset developed based on the Charades dataset~\cite{sigurdsson2016hollywood} by adding sentence temporal annotations. It contains 3720 video-query pairs for testing.

\subsection{Main Results}
The main results of video grounding for LLMs are shown in Table~\ref{tab:total} and Table~\ref{tab:VidLLM}. We summarize our findings as follows.

\noindent\textbf{Observation 1. LLMs show preliminary abilities for video grounding tasks, outperforming the VidLLMs. }

On the one hand, all the VidLLMs we test are not as good as the random method in completing the video grounding tasks, which indicates that the current VidLLMs are still far from satisfying video temporal understanding, and more temporal-related video tasks should be added to further finetune these model.

On the other hand, although some combination methods of LLMs and visual model we tried cannot outperform the random method, on average, GPT-3.5 has shown better performance improvement over the VidLLMs and random results,  indicating that LLMs are indeed able to understand the visual description and questions for video grounding and use the corresponding temporal information to provide reasonable answers. Our combination of LLMs and visual models has been proven to be effective. Overall, we can find that LLMs have the ability of temporal information understanding. 

\noindent\textbf{Observation 2. Different combinations of visual description generators, LLMs, and prompt designs, can significantly affect the recall rate of video grounding}

As shown in Table~\ref{tab:total}, using the same valid rate as the evaluation metrics, we can see a huge difference in video grounding performance when changing the visual description generators, LLMs, and prompt designs. For example, compared with Vicuna-7B using zero-shot prompts and the Fc model, GPT-3.5 using the VQA model of Blip and one-shot with confidence judgment prompts has a significantly better performance in conducting video grounding tasks (from 25.83 to 33.87, with a performance difference of more than 30\%). However, considering that the current video grounding model has higher performance in the same dataset (for instance, R@1 Iou=0.5 could achieve more than 40~\cite{zeng2020dense}), these results show that it is worth further studying and analyzing the impact of different models and methods to better reveal their impact on the results of video grounding using LLMs for achieving higher performance.

\subsection{Results with Different LLMs}
We compared different LLMs, including GPT-3.5, Vicuna-7B, Longchat-7B, and some VidLLMs such as Video-Chat, Video-ChatGPT, and Video-LLaMA. For normal LLMs, we used the aforementioned combination of LLMs and visual descriptions to complete the video grounding task. For VidLLMs, we directly asked them to read the corresponding video content and answer the video grounding question. The final results are shown in Table~\ref{tab:llm}, and we can draw the following conclusion based on this.

\noindent\textbf{Observation 3. LLMs' ability to complete video grounding tasks not only depends on the model scale but is also related to the models' ability of handling long sequence question answers.}

As shown in Table~\ref{tab:total} and~\ref{tab:llm}, we can clearly see that GPT-3.5 achieves higher results in video grounding tasks than Vicuna-7B and Longchat-7B in most cases, indicating that larger LLM can perform better in video grounding tasks.

As for the performance difference between small-size LLMs, although Longchat and Vicuna are both finetuned from LLaMA~\cite{touvron2023llama}, we can see from the table that in most cases (i.e. using different visual models and prompt methods), Longchat-7B shows better results than Vicuna-7B in video grounding under the same conditions (Vicuna-7B is even worse than random results in many circumstances). The main reason may be that the prompt we input for video grounding usually has thousands or even nearly 10000 tokens, while the Longchat-7B model has extra condensing rotary embeddings and finetuning for long-context data, showing better long-context capability than Vicuna.

\begin{table}[!htbp]
\small
    \centering
    \begin{tabular}{ccccc}
    \toprule
        Valid Rate & IoU=0.3 & IoU=0.5& IoU=0.7  \\
    \midrule
        GPT-3.5& 33.87& 11.80 & 4.22\\
        Vicuna-7B & 23.63 &9.76 &3.15  \\
        Longchat-7B&24.57&10.91&3.92\\
        Video-Chat &9.03&3.31&1.26     \\
        Video-ChatGPT&20.00&7.69&1.75\\
        Video-LLaMA&10.38&3.84&0.91\\
    \midrule
    Random&23.26&9.06&2.88\\
    \bottomrule
    \end{tabular}
    \caption{The overall best performance on the video grounding with different LLMs.} 
    \vspace{-0.6cm}
    \label{tab:llm}
\end{table}

\subsection{Results with Different Visual Models }

We compare the results in completing video grounding tasks when receiving visual descriptions generated by different generators, and we have observations as follows:

\noindent\textbf{Observation 4. General advanced caption models as visual description models do not guarantee a performance boost in helping LLMs conduct video grounding tasks.}

As shown in Table~\ref{tab:total} and~\ref{tab:generator}, some advanced caption methods, such as Blip, as visual description generators with higher CIDEr value, may achieve higher performance than other methods in a few evaluation metrics. Generally, it can increase the number of cases with $IoU\geq0.3$, which shows that a more efficient caption description of video content, to some extent, can better activate LLMs' ability to capture key content and conduct spatiotemporal reasoning. However, no caption methods as visual models consistently achieve the best performance on all the evaluation metrics, even for the well-known model that obtained high performance of image caption. This result calls for the need to design a more fine-grained caption model to describe the video content second by second in detail.
\begin{table}[h]
\small
    \centering
    \begin{tabular}{ccccc}
    \toprule
        Valid Rate & IoU=0.3 & IoU=0.5& IoU=0.7&CIDEr  \\
        \midrule
        Fc&25.83&9.68&3.17&1.053	\\
        Att2in&28.23&11.34&3.17&1.195\\
        Transformer&26.32&10.62&3.25&1.303\\
        Updown&25.13&10.91&3.92&1.239\\
        Blip&30.67&11.26&3.84&1.335\\
        Blip(VQA)&33.87&11.80&4.22&-\\
    \bottomrule
    \end{tabular}
    \vspace{-0.2cm}
    \caption{The overall best performance on the video grounding with different visual models. CIDEr represents a metric for measuring models' captioning capability~\cite{vedantam2015cider}. Generally speaking, caption models with higher CIDEr values are supposed to exhibit better caption capabilities.}
    \vspace{-0.4cm}
    \label{tab:generator}
\end{table}

\noindent\textbf{Observation 5. Introducing additional query information into the description of video content can significantly improve the ability of LLMs to conduct video grounding, even with a small amount of additional information.}

In the process of designing visual description generators, although we mostly use caption-based generators for descriptions, they occasionally fail to include keywords of grounding query. Thereby, we also introduce the VQA-based generator to enhance the description, since the answer to \textit{`Is it currently happening $<$query event$>$ in the image'} is strongly related to the \textit{$<$query event$>$}, bringing extra information. Through our observation, with the addition of VQA information, the LLMs have achieved improvements in most metrics of video grounding, indicating that we still have significant potential for improvement in visual description generator design by introducing incremental information.

        

\subsection{Results with Different Prompting Methods}

As shown in Table~\ref{tab:total} and ~\ref{tab:prompt}, we make comparisons with different prompting methods, which include zero-shot prompting and one-shot prompting. Compared to the zero-shot method, the one-shot prompting method additionally adds exemplars for video grounding tasks. We can see that when the one-shot prompt method and the zero-shot method using different visual models are input to different LLMs, the presence or absence of example can not decisively improve the effect of video grounding, and different LLMs show different degrees of sensitivity to them.

In order to further explore whether the prompt design can help LLMs complete the video grounding task more effectively, and in response to the situation where visual description information sometimes appears vague and cannot accurately guide LLMs and humans in video grounding, we design a judgment guidance prompt that allows LLMs to consider whether the given information is suitable for video grounding before the prediction, and we can have the following observations.

\begin{table}[h]
    \centering
    \begin{tabular}{ccccc}
    \toprule
        Valid Rate & IoU=0.3 &  IoU=0.5 &  IoU=0.7 \\
    \midrule
        Zero-shot& 25.97&10.91 &3.47  \\
        One-shot & 26.02 &10.91 &3.04  \\
    \midrule
        \makecell{One-shot \\Confidence judgment} &  33.87  &11.80  &4.22   \\
    \bottomrule
    \end{tabular}
    \caption{The overall best performance on the video grounding with different prompting methods.} 
    \vspace{-0.6cm}
    \label{tab:prompt}
\end{table}



\noindent\textbf{Observation 6. The prompting method of instructing LLMs to separately judge the predictability and infer results can significantly improve the performance of video grounding.}

As shown in Table~\ref{tab:prompt}, comparing different one-shot prompt methods, there are only a few differences between the two methods (i.e., the one-shot with confidence judgment prompt has an additional sentence \textit{`judge whether the description sequence is suitable for the video grounding'}), which leads to a significant change in the prediction effect of video grounding, and the confidence gadget significantly improves the prediction recall, from 26.02 to 33.87. We analyze the reasons for the improvement, which may stem from LLMs directly giving answers of \textit{`unpredictable'} in some cases with low predictability. However, this does not mean that the description sequence generated by the visual model is completely unable to be used for video grounding tasks. In this case, our prompt with confidence judgment can better instruct LLMs to infer video grounding results based on existing information (and we will provide similar examples in the following case analysis section).


\begin{figure*}[h]
    \centering
    \includegraphics[width=0.95\linewidth]{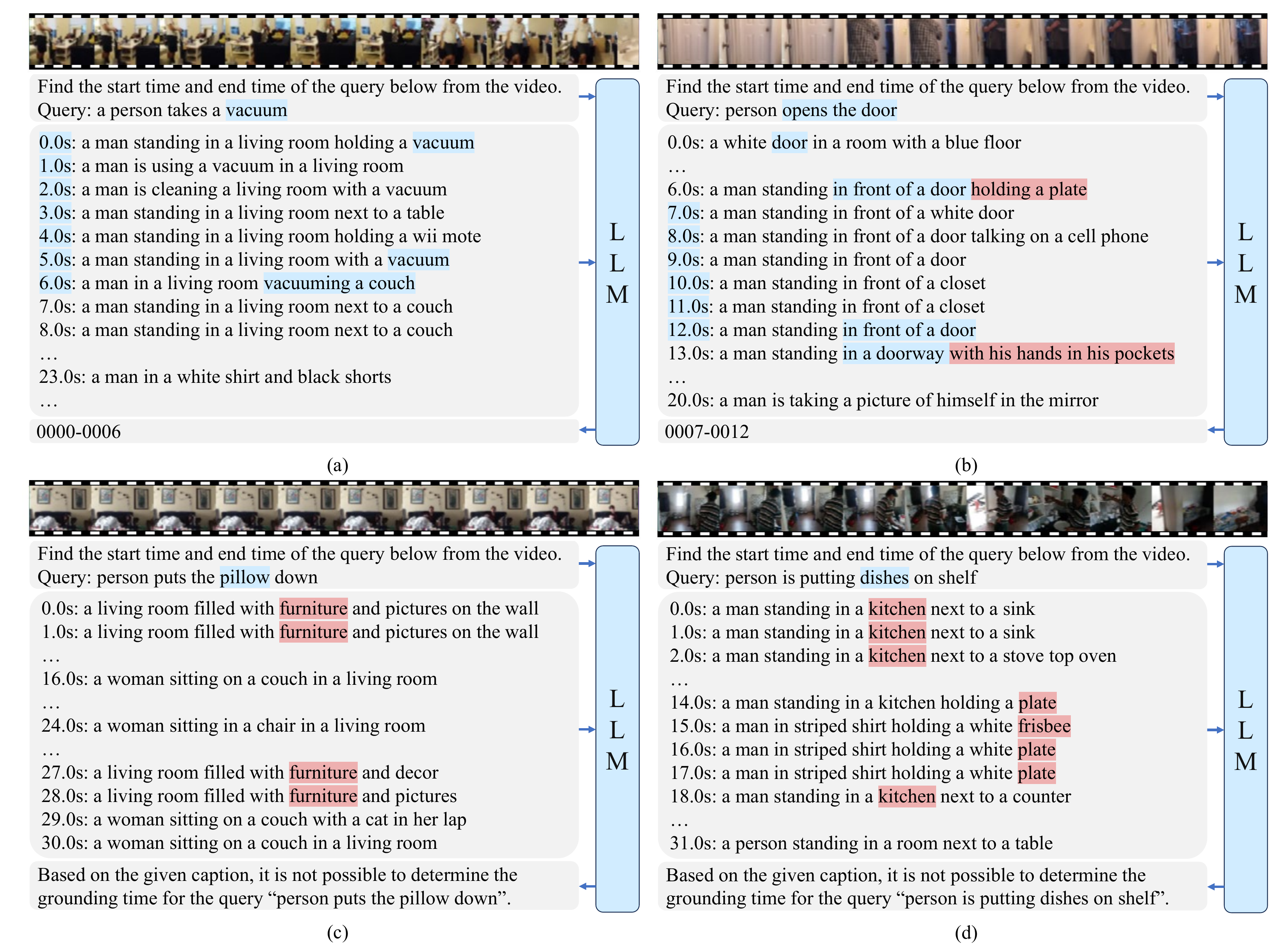}
    \vspace{-0.3cm}
    \caption{Example cases of LLMs conducting video grounding task, (a) and (b) are successful cases, while (c) and (d) are failure cases, since LLMs give the answer \textit{`Based on the given caption, it is not possible to determine the grounding time for the query'}. The text with a \textbf{blue} background represents \textbf{positive} for grounding answers, while the text with a \textbf{red} background represents \textbf{negative} for grounding answers, although it might be related to the query.}
    \vspace{-0.4cm}
    \label{fig:case}
\end{figure*}

\subsection{Examples and Case Analysis}
In Figure~\ref{fig:case}, we present four prompt examples for LLMs to complete video rounding, as shown in Figure~\ref{fig:case}, including successful and failed examples. Based on these actual cases, we can draw the following observations.

\noindent\textbf{Observation 7. LLMs infer from the actually received information and complete the video grounding task, rather than randomly guessing}

In Figure~\ref{fig:case}(a), the visual description for the video directly mentions the keyword \textit{`vacuum'} in the initial description sequence, ensuring that LLMs can easily infer when \textit{`a person takes a vacuum'} occurs; In Figure~\ref{fig:case}(b), although the visual description sequence did not directly mention the action of \textit{`open the door'}, it concentrates on mentioning the word-door for a period of time, which helps LLMs effectively infer the start and end time of the video grounding task based on the occurrence of \textit{`a man'} in the description. These successful cases demonstrate that LLMs have the ability to infer video grounding answers based on corresponding textual information while generating effective visual descriptions for videos. However, in some cases, LLMs are unable to complete the visual grounding task according to the description of the visual description sequence for different reasons, and LLMs would respond \textit{` it's not possible to determine the grounding time of the query'}. These failed cases and successful cases prove that LLMs are indeed trying to infer the answer of video grounding instead of randomly guessing.

\noindent\textbf{Observation 8. The reason for the failure case is mainly from the vague description of the visual models, and the secondary one is the insufficient reasoning ability of LLMs in the case of weak information.}

The reasons for the failure cases of video grounding mainly lie in the incomplete visual description and the key information not mentioned. As shown in Figure~\ref{fig:case}(c), due to the lack of keyword \textit{`pillow'} mentioned in the visual description sequence, LLMs cannot effectively confirm the start and end time of the query \textit{`person puts the pilot down'}. On the other hand, due to the fuzziness of the description for the video generated by the caption model, it will increase the reasoning difficulty for LLMs. As shown in Figure~\ref{fig:case}(d), although the \textit{`plate'} and \textit{`kitchen'} mentioned in the description sequence can be seen to be highly correlated with the \textit{`dishes'} mentioned in the query, LLMs still gave the answer \textit{`impossible for grounding'}, which shows that the reasoning ability of LLMs in the case of weak information still need to be strengthened.

\vspace{-0.2cm}
\section{Conclusion}
\vspace{-0.2cm}
In this paper, we propose LLM4VG, a comprehensive benchmark that systematically evaluates the performance of different LLMs on the video grounding task, with our proposed combination prompting method of various visual models and LLMs for video grounding. We evaluate and analyze the performances using different visual models, LLMs, and prompting methods. Our evaluation results demonstrate that the existing VidLLMs are still far from satisfying video temporal understanding, requiring temporal training tasks. The combination of visual models and LLMs shows preliminary abilities for video grounding tasks, achieving higher performance than VidLLMs. We conclude that more fine-grained visual models and prompting methods with further guidance of instructions are required to help LLMs better conduct video grounding tasks.



{
    \small
    \bibliographystyle{ieeenat_fullname}
    \bibliography{main}
}
\section{Supplementary matierals}
As shown in Table~\ref{tab:dataset}, we provide the extra experiment of LLMs' video grounding performance on the ActivityNet-Captions dataset~\cite{krishna2017dense}, which shares many similar trends compared to the Charades-STA dataset.

\begin{table}[!htbp]
\scriptsize
    \centering
    \caption{Performance of video grounding on ActivityNet-Captions dataset with different Large Language Models. } 
    \begin{tabular}{ccccccc}
    \toprule
        Valid Rate&IoU=0.3&IoU=0.5&IoU=0.7\\
    \midrule
        Video-Chat&8.8&3.7&1.5\\
        Video-LLaMA&6.9&2.1&0.8\\
        Video-ChatGPT&26.4&13.6&6.1\\
        Vicuna& 17.37 &8.26 &2.94\\
        Longchat&19.78&9.45&3.35\\
        GPT-3.5&33.51& 14.97 & 7.43\\

    \bottomrule
    \end{tabular}
    
    \label{tab:dataset}
    
\end{table}


\end{document}